\documentclass{article}
\usepackage{spconf,amsmath,graphicx}
\usepackage{amssymb}

\usepackage{enumitem}
\setlist{nosep, leftmargin=14pt}

\usepackage{multirow}
\usepackage{booktabs} 
\usepackage{makecell}

\usepackage{mwe} 

\title{Personalized Federated Learning with Residual Fisher Information for Medical Image Segmentation}

\name{Meilu Zhu, Yuxing Li, Zhiwei Wang, Edmund Y. Lam
\thanks{Corresponding author: Edmund Y. Lam (elam@eee.hku.hk)}
\thanks{This work is supported by the Innovation and Technology Fund (ITS/341/23) of Hong Kong, China.}
}
\address{Department of Electrical and Electronic Engineering, The University of Hong Kong, Hong Kong, China}
\begin{document}
%
\maketitle

Federated learning enables multiple clients (institutions) to collaboratively train machine learning models without sharing their private data. To address the challenge of data heterogeneity across clients, personalized federated learning (pFL) aims to learn customized models for each client. In this work, we propose pFL-ResFIM, a novel pFL framework that achieves client-adaptive personalization at the parameter level. Specifically, we introduce a new metric, Residual Fisher Information Matrix (ResFIM), to quantify the sensitivity of model parameters to domain discrepancies. To estimate ResFIM for each client model under privacy constraints, we employ a spectral transfer strategy that generates simulated data reflecting the domain styles of different clients. Based on the estimated ResFIM, we partition model parameters into domain-sensitive and domain-invariant components. A personalized model for each client is then constructed by aggregating only the domain-invariant parameters on the server. Extensive experiments on public datasets demonstrate that pFL-ResFIM consistently outperforms state-of-the-art methods, validating its effectiveness.

\section{Introduction}

Federated Learning (FL) allows multiple clients (hospitals) to collaboratively learn a shared model while keeping data localized — making it particularly valuable for privacy-sensitive medical scenarios~\cite{zhu2024deer}. However, the performance of the standard FL framework (i.e., FedAvg~\cite{fedavg}) is often compromised by statistical heterogeneity across clients. When each client learns from distinct data distributions, their local updates may conflict, causing the aggregated global model to perform poorly~\cite{zhu2025fedbm,mao2022fedaar,zhu2024stealing,zhu2023fedoss}.

Personalized federated learning (pFL) has emerged as an alternative framework that enables selective knowledge transfer across the federation~\cite{chen2021personalized}. pFL produces separate models tailored to each client to adapt to local data, rather than a shared global model. One representative type of pFL algorithm is decoupling the parameters of a FL model into the domain-invariant and domain-sensitive subnetworks and keeping the latter locally to avoid inter-client interference. For example, FedBN~\cite{li2021fedbn} and SioBN~\cite{andreux2020siloed} consider batch normalization layers to be domain-sensitive and keep them updated locally, without communicating and aggregating them at the server. In addition, FedPer~\cite{arivazhagan2019federated} only aggregates feature extractors because they are less affected by domain discrepancy. However, these methods achieve limited performance due to two issues. (1) Layer-level selection strategy is empirically-driven and sub-optimal, and cannot balance inter-client knowledge sharing and local personalization. (2) As the model training procedure is dynamic, the client-specific subnetwork should vary across optimization iterations.

To address these problems, we introduce pFL-ResFIM, a new personalized federated learning framework based on Residual Fisher Information Matrix (ResFIM), which aims to achieve client-adaptive and atom-level personalization. Specifically, we first introduce a ResFIM metric to measure the sensitivity of model parameters to domain discrepancy at the parameter level. To estimate the ResFIM of each client model in FL setting without compromising data privacy, we propose to leverage a spectral transfer based strategy generate simulated data with domain styles of different clients. Based on the estimated ResFIM, the parameters of each client model can be separated into domain-sensitive and domain-invariant components. A personalized model for each client is then constructed by aggregating only the domain-invariant parameters on the server.
The contributions of this work are summarized as follows: \textbf{(1)} We propose pFL-ResFIM, a new personalized FL framework that utilizes the Fisher Information Matrix to achieve client-adaptive and atom-level personalization. \textbf{(2)} We introduce a ResFIM metric that effectively measure the sensitivity of client model parameters to domain discrepancies among clients. \textbf{(3)} We conduct extensive experiments on public datasets to evaluate pFL-ResFIM. The results show the superior performance of pFL-ResFIM against the state-of-the-art approaches.

\section{Related Work}
 Personalized Federated Learning (pFL) has attracted wide attention for its potential to address data heterogeneity. Existing pFL methods can be broadly classified into three main types.

The first category focuses on deriving client-specific aggregation weights to tailor the global model for each client, thereby mitigating interference among local models. For instance, FedAS~\cite{yang2024fedas} employs the Fisher Information Matrix (FIM) of local models to determine aggregation weights, while pFedLA~\cite{Ma_2022_CVPR} directly optimizes personalized weighting schemes at the server. The second category uses knowledge distillation to facilitate knowledge transfer among heterogeneous client models. Methods such as FedDF~\cite{lin2020ensemble} utilize ensemble distillation to integrate knowledge from multiple local models, whereas FML~\cite{shen2020federated} adopts a mutual learning strategy, enabling clients to collaboratively train a generalized model while independently refining their personalized counterparts. The third category involves decomposing model parameters into domain-invariant and domain-sensitive components. In this paradigm, domain-invariant parameters are aggregated to promote knowledge sharing, while domain-sensitive parameters are retained locally to preserve personalization. For example, FedBN~\cite{li2021fedbn} and FedPer~\cite{arivazhagan2019federated} identify batch normalization layers and classifier layers, respectively, as domain-sensitive components. 

While these methods have shown empirical success, they are largely heuristic-driven, static, and operate at a layer-wise granularity. In contrast, this work aims to explore adaptive parameter-level personalization, offering a more fine-grained and dynamic approach to model customization.

\section{Methodology}

In this section, we present our pFL-ResFIM, which aims to achieve personalized federated learning guided by Residual Fisher Information Matrix (ResFIM). The core idea of pFL-ResFIM is to use ResFIM to identify domain-sensitive parameters of client models. We propose to preserve these parameters at the client side for model personalization and only aggregate the remaining for knowledge sharing across clients.

\begin{figure}[!t]
    \centering
    \includegraphics[width=\linewidth]{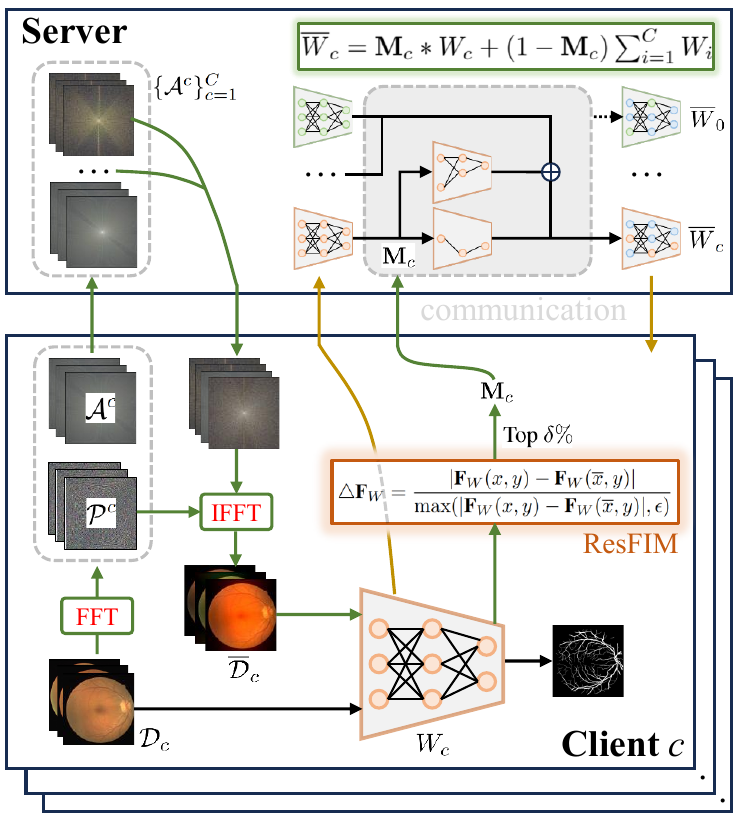}
    \vspace{-7mm}
    \caption{The overall architecture of the proposed pFL-ResFIM framework.}
    \label{fig:framework}
\end{figure}

\subsection{Residual Fisher Information Matrix}

Fisher Information Matrix (FIM) is a cornerstone of statistical estimation theory, quantifying the amount of information an observable random variable carries about an unknown parameter. In deep learning, FIM is adopted to capture the sensitivity of the model's output distribution to its parameters, which asymptotically approximates the second-order curvature of the loss landscape around a given point. Formally, for a model with parameters $W$, FIM is defined as:
\vspace{-1.0mm}
\begin{equation}
\vspace{-1.0mm}
\textbf{F}_{W} = \mathbb{E}_x [\mathbb{E}_{y\sim f_W(y|x)}\bigtriangledown_W\mathcal{L}(f_W(x), y)\cdot \bigtriangledown_W\mathcal{L}(f_W(x), y)^\top],
\label{eq1}
\nonumber
\end{equation}
where $x$ is a sample with the label $y$. $\textbf{F}_{W} \in \mathbb{R}^{|W|\times|W|}$ is the symmetrical matrix and can be efficiently approximated via the diagonal approximation to reduce the computational overhead~\cite{matena2022merging}. $\bigtriangledown_W\mathcal{L}(\cdot)$ represents the gradient of the loss function with respect to the parameters $W$. Intuitively, $\textbf{F}_{W}$ captures how much changing the parameters affects the model's output, thereby measuring the importance of each parameter.

However, FIM can not sufficiently capture the behavior of parameters in cross-domain scenes, particularly in FL, where parameters can exhibit significantly different sensitivities to varying data distributions. Consequently, a more nuanced metric is required to quantify the variation of parameters across different domains. To this end, we propose to calculate the Residual Fisher Information Matrix (ResFIM), denoted as $\bigtriangleup \textbf{F}_{W}$, to directly measure parameter sensitivity to domain discrepancies. It is defined as:
\vspace{-2.0mm}
\begin{equation}
\vspace{-1.0mm}
\bigtriangleup \textbf{F}_{W} = \frac{|\textbf{F}_{W}(x, y) - \textbf{F}_{W}(\overline{x}, y)|}{\max(|\textbf{F}_{W}(x, y) - \textbf{F}_{W}(\overline{x}, y)|, \epsilon)},
\label{eq2}
\end{equation}
where $\epsilon$ is a small constant to prevent division by zero. $x$ and $\overline{x}$ are samples from different domains. $|\textbf{F}_{W}(x, y) - \textbf{F}_{W}(\overline{x}, y)|$ computes the difference between the FIM across domains, reflecting the model's varying sensitivity to parameter changes in different data distributions. A higher $\bigtriangleup \textbf{F}_{W(p)}$ indicates that the parameter $W(p)$ is more sensitive to domain changes.

\begin{table*}[!t]
\vspace{-2mm}
\caption{The Dice performance of different methods on the prostate dataset.}
\center
\renewcommand\arraystretch{1}
\setlength{\tabcolsep}{0pt}
\begin{tabular}{p{60pt}|p{60pt}|p{60pt}|p{60pt}|p{60pt}|p{60pt}|p{60pt}|p{60pt}}
\toprule[1pt]
\makecell[c]{Methods} & \makecell[c]{BIDMC} & \makecell[c]{BMC} & \makecell[c]{HK} & \makecell[c]{I2CVB} & \makecell[c]{RUNMC} & \makecell[c]{UCL} & \makecell[c]{Average}\\
 \hline
\makecell[c]{FedAvg~\cite{fedavg}}  & \makecell[c]{$73.18\pm0.70$} & \makecell[c]{$86.34\pm0.74$} & \makecell[c]{$86.32\pm1.46$} & \makecell[c]{$87.48\pm1.11$} & \makecell[c]{$88.59\pm0.71$} & \makecell[c]{$79.34\pm3.53$} & \makecell[c]{$83.54\pm5.50$} \\ 
\makecell[c]{FedBN~\cite{li2021fedbn}}   & \makecell[c]{$77.56\pm4.45$} & \makecell[c]{$\textbf{88.45}\pm0.83$} & \makecell[c]{$86.25\pm1.65$} & \makecell[c]{$\textbf{87.77}\pm1.73$} & \makecell[c]{$87.55\pm0.40$} & \makecell[c]{$79.68\pm2.56$} & \makecell[c]{$84.54\pm4.28$} \\

\makecell[c]{SioBN~\cite{andreux2020siloed}}   & \makecell[c]{$71.47\pm5.19$} & \makecell[c]{$87.33\pm0.90$} & \makecell[c]{$88.22\pm0.51$} & \makecell[c]{$86.60\pm2.46$} & \makecell[c]{$87.14\pm0.50$} & \makecell[c]{$80.49\pm1.47$} & \makecell[c]{$83.54\pm5.96$} \\

\makecell[c]{FedPer~\cite{arivazhagan2019federated}}  & \makecell[c]{$72.10\pm5.37$} & \makecell[c]{$86.98\pm1.38$} & \makecell[c]{$85.26\pm3.78$} & \makecell[c]{$86.83\pm1.36$} & \makecell[c]{$88.30\pm0.68$} & \makecell[c]{$81.06\pm3.96$} & \makecell[c]{$83.42\pm5.55$} \\
 \hline
\makecell[c]{pFL-ResFIM}  & \makecell[c]{$\textbf{78.07}\pm3.30$} & \makecell[c]{$88.25\pm0.98$} & \makecell[c]{$\textbf{88.69}\pm1.25$} & \makecell[c]{$87.59\pm2.61$} & \makecell[c]{$\textbf{88.82}\pm0.58$} & \makecell[c]{$\textbf{81.09}\pm2.01$} & \makecell[c]{$\textbf{85.42}\pm4.23$}\\
\bottomrule[1pt]
\end{tabular}
\label{tab1}
\end{table*}

\begin{table*}[!t]
\vspace{-5mm}
\caption{The Dice performance of different methods on the vessel dataset.}
\center
\renewcommand\arraystretch{1}
\setlength{\tabcolsep}{0pt}
\begin{tabular}{p{60pt}|p{60pt}|p{60pt}|p{60pt}|p{60pt}|p{60pt}|p{60pt}|p{60pt}}
\toprule[1pt]
 \makecell[c]{Methods} & \makecell[c]{ChaseDB} & \makecell[c]{DR-Hagis} & \makecell[c]{DRIVE} & \makecell[c]{HRF} & \makecell[c]{LES-AV} & \makecell[c]{ORVS} & \makecell[c]{Average}  \\
\hline
\makecell[c]{FedAvg~\cite{fedavg}}  & \makecell[c]{$68.24\pm0.74$} & \makecell[c]{$69.80\pm0.10$} & \makecell[c]{$76.06\pm0.07$} & \makecell[c]{$72.51\pm0.09$} & \makecell[c]{$73.30\pm0.06$} & \makecell[c]{$71.56\pm0.02$} & \makecell[c]{$71.91\pm2.49$} \\

\makecell[c]{FedBN~\cite{li2021fedbn}}  & \makecell[c]{$69.67\pm0.83$} & \makecell[c]{$70.80\pm0.02$} & \makecell[c]{$76.16\pm0.08$} & \makecell[c]{$71.35\pm0.17$} & \makecell[c]{$72.83\pm0.38$} & \makecell[c]{$71.60\pm0.07$} & \makecell[c]{$72.07\pm2.05$} \\

\makecell[c]{SioBN~\cite{andreux2020siloed}}   & \makecell[c]{$68.01\pm0.21$} & \makecell[c]{$70.34\pm0.15$} & \makecell[c]{$75.55\pm0.10$} & \makecell[c]{$70.89\pm0.16$} & \makecell[c]{$73.49\pm0.69$} & \makecell[c]{$\textbf{71.64}\pm0.09$} & \makecell[c]{$71.65\pm2.38$} \\

\makecell[c]{FedPer~\cite{arivazhagan2019federated}} & \makecell[c]{$70.80\pm0.17$} & \makecell[c]{$69.45\pm0.18$} & \makecell[c]{$75.90\pm0.02$} & \makecell[c]{$73.00\pm0.01$} & \makecell[c]{$73.61\pm0.26$} & \makecell[c]{$71.14\pm0.16$} & \makecell[c]{$72.32\pm2.11$} \\


 \hline
\makecell[c]{pFL-ResFIM}  & \makecell[c]{$\textbf{72.74}\pm0.25$} & \makecell[c]{$\textbf{71.63}\pm0.17$} & \makecell[c]{$\textbf{76.20}\pm0.13$} & \makecell[c]{$\textbf{73.31}\pm0.06$} & \makecell[c]{$\textbf{74.58}\pm0.28$} & \makecell[c]{$71.50\pm0.13$} & \makecell[c]{$\textbf{73.33}\pm1.65$} \\

\bottomrule[1pt]
\end{tabular}
\label{tab2}
\end{table*}

\subsection{Personalized Federated Learning with ResFIM}
Next, we describe how to integrate ResFIM into a FL system to achieve client-adaptive and atom-level personalization.

\noindent\textbf{Overview of pFL-ResFIM} We consider a pFL system comprising a central server and $C$ clients, collaboratively training a set of personalized models $\{W_c\}^C_{c=1}$. Each client $c$ possesses a private dataset $\mathcal{D}_c = \{(x^c_i,y^c_i)\}^{N_c}_{i=1}$ with $N_c$ samples. The objective the system is formulated as:
\vspace{-3mm}
\begin{equation}
\vspace{-2mm}
\min_{W_1,...,W_C} f(W_1,...,W_C) = \frac{1}{C} \sum_{c=1}^C \mathbb{E}_{(x^c_i,y^c_i)\thicksim \mathcal{D}_c} \mathcal{L}(W_c, x^c_i).
\label{eq3}
\nonumber
\end{equation}
The overall optimization process proceeds through communication between clients and the server for multiple rounds, as shown in Fig.~\ref{fig:framework}. In each round, every client $c$ first performs local training on $W_c$ using its private data $\mathcal{D}_c$. Next, the client estimates the ResFIM $\bigtriangleup \textbf{F}_{W_c}$ for its local model and generates a binary mask $\mathbf{M}_c$ by selecting the top $\delta\%$ of parameters with the largest values in $\bigtriangleup \textbf{F}_{W_c}$. The local models $\{W_c\}^C_{c=1}$ and their corresponding masks $\{\mathbf{M}_c\}^C_{c=1}$ are then transmitted to the server. The server proceeds to construct a personalized model $\overline{W}_c$ for each client via the following aggregation:
\vspace{-2mm}
\begin{equation}
\vspace{-2mm}
\overline{W}_c= \mathbf{M}_c*W_c + (1-\mathbf{M}_c)\sum_{i=1}^C W_i.
\label{eq4}
\end{equation}
Here, the first term on the right side preserves the client's domain-specific knowledge by retaining its sensitive parameters locally. The second term achieves knowledge sharing across the federation by aggregating the domain-invariant parameters of all clients.

\noindent\textbf{Estimation of ResFIM in FL} As shown in Eq.~(\ref{eq2}), computing the ResFIM $\bigtriangleup \textbf{F}_{W_c}$ for a client $c$ requires data from all other clients $\{\mathcal{D}_j\}_{j=1}^{C\setminus c}$. Direct data sharing is infeasible due to strict privacy constraints. To simulate the data of other clients with privacy leakage, we leverage a spectral transfer based method inspired by \cite{yang2020fda,liu2021feddg}. Specifically, for an image $x_i^c \in \mathbb{R}^{H \times W \times D}$ of the client $c$, we first apply the Fast Fourier Transform (FFT) to project it into the frequency domain, obtaining the signal $\mathcal{F}(x_i^c)$: 
$\mathcal{F}(x_i^c)(u,v,d) = \sum_{h=0}^{H-1}\sum_{w=0}^{W-1} x_i^c(h,w,d)e^{-j2\pi(\frac{h}{H}u+\frac{w}{W}v)}.$
$\mathcal{F}(x_i^c)$ can be further decomposed into an amplitude spectrum $\mathcal{A}_i^c$ and a phase spectrum $\mathcal{P}_i^c$, where the low-frequency components of $\mathcal{A}_i^c$ primarily encapsulate the domain style.
Before federated training begins, all clients upload their amplitude spectrum sets $\{\mathcal{A}^c\}_{c=1}^C \in \mathbb{R}^{N\times H\times W\times D}$ to the server in a one-time initialization step, where $N$ is the total sample size of clients and $\mathcal{A}^c = \{\mathcal{A}_i^c\}_{i=1}^{N_c}$. 
The server then distributes the set $\{\mathcal{A}^j\}_{j=1}^{C\setminus c}$ to each client $c$. During local training, for each image $x_i^c$, the client $c$ randomly samples an amplitude spectrum from each $\mathcal{A}^j$ in $\{\mathcal{A}^j\}_{j=1}^{C\setminus c}$. 
It then replaces the low-frequency component of its own $\mathcal{A}_i^c$ with the low-frequency component from the sampled spectra. These modified amplitude spectra are combined with the original phase spectrum $\mathcal{P}_i^c$ and transformed back via inverse FFT, generating a set of simulated images $\{\overline{x}_{i,k}^c\}_{k=1}^{C-1}$. This process creates a simulated dataset $\overline{\mathcal{D}}_c = \{(\overline{x}^c_i,y^c_i)\}^{N_c*(C-1)}_{i=1}$ that preserves the semantic content of $\mathcal{D}_c$ while exhibiting the style variations of other clients. The ResFIM $\bigtriangleup \textbf{F}_{W_c}$ can then be reliably estimated using the local data $\mathcal{D}_c$ and the simulated data $\overline{\mathcal{D}}_c$.

\section{Experiments}

\subsection{Datasets and Implementation Details}

\textbf{Datasets}. \textbf{(1)} \textsc{Prostate Segmentation}. This dataset contains prostate T2-weighted MRI data of 116 patients collected from six institutions~\cite{liu2021feddg}. The samples of each client are divided into training, validation and test sets as 6:2:2. All 3D volumes are sliced into images with the axial plane. \textbf{(2)} \textsc{Vessel Segmentation}. Retina images are collected from ChaseDB~\cite{ChaseDB}, DR-Hagis~\cite{DR-Hagis}, DRIVE~\cite{staal2004ridge}, HRF~\cite{HRF}, LES-AV~\cite{LES-AV}, ORVS~\cite{sarhan2021transfer}, forming six clients. Based on original partition, their training and sets contain 20 and 8, 34 and 6, 20 and 20, 15 and 30, 17 and 5, 41 and 8 images, respectively. \textbf{(3)} \textsc{Meibomian Gland Segmentation}. The infrared images of this dataset are from two clients. Their data are collected from the Lo Fong Shiu Po Eye Centre (LFSC) in Hong Kong and a public dataset MGD-1K~\cite{saha2022automated}, respectively. The partition way is same with the prostate dataset.

\noindent \textbf{Implementation}. The proposed pFL-ResFIM and all baseline methods are implemented with PyTorch library. We adopt UNet as the backbone network of all methods. For all three datasets, we use the Adam optimizer with a batch size of 4. The default numbers of local epochs and communication rounds are 4 and 100. The prostate dataset uses a fixed learning rate of $1\times10^{-4}$ while the vessel and gland datasets use a initial learning rate of $1\times10^{-3}$ with a decaying rate of 0.99 per round. Model performance is evaluated using the Dice similarity coefficient. For each dataset, we conduct three trials and present the mean and the standard deviation.

\begin{table}[!t]
\caption{Comparison of various methods on gland dataset.}
\center
\renewcommand\arraystretch{1}
\setlength{\tabcolsep}{0pt}
\begin{tabular}{p{60pt}|p{60pt}|p{60pt}|p{60pt}}
\toprule[1pt]
 \makecell[c]{Methods} & \makecell[c]{LFSC} & \makecell[c]{MGD-1K} & \makecell[c]{Average}  \\
\hline
\makecell[c]{FedAvg~\cite{fedavg}}  & \makecell[c]{$74.52\pm1.86$} & \makecell[c]{$81.86\pm0.96$} & \makecell[c]{$78.19\pm3.67$} \\

\makecell[c]{FedBN~\cite{li2021fedbn}}  & \makecell[c]{$74.18\pm3.26$} & \makecell[c]{$81.86\pm1.68$} & \makecell[c]{$78.02\pm3.99$} \\

\makecell[c]{SioBN~\cite{andreux2020siloed}}  & \makecell[c]{$63.42\pm7.04$} & \makecell[c]{$81.49\pm0.76$} & \makecell[c]{$72.45\pm9.03$} \\

\makecell[c]{FedPer~\cite{arivazhagan2019federated}} & \makecell[c]{$72.67\pm3.08$} & \makecell[c]{$81.72\pm2.00$} & \makecell[c]{$77.20\pm4.52$} \\

 \hline
\makecell[c]{pFL-ResFIM}  & \makecell[c]{$\textbf{76.58}\pm1.75$} & \makecell[c]{$\textbf{82.32}\pm1.01$} & \makecell[c]{$\textbf{79.45}\pm2.86$} \\

\bottomrule[1pt]
\end{tabular}
\label{tab3}
\end{table}

\begin{figure}
    \centering
    \includegraphics[width=\linewidth]{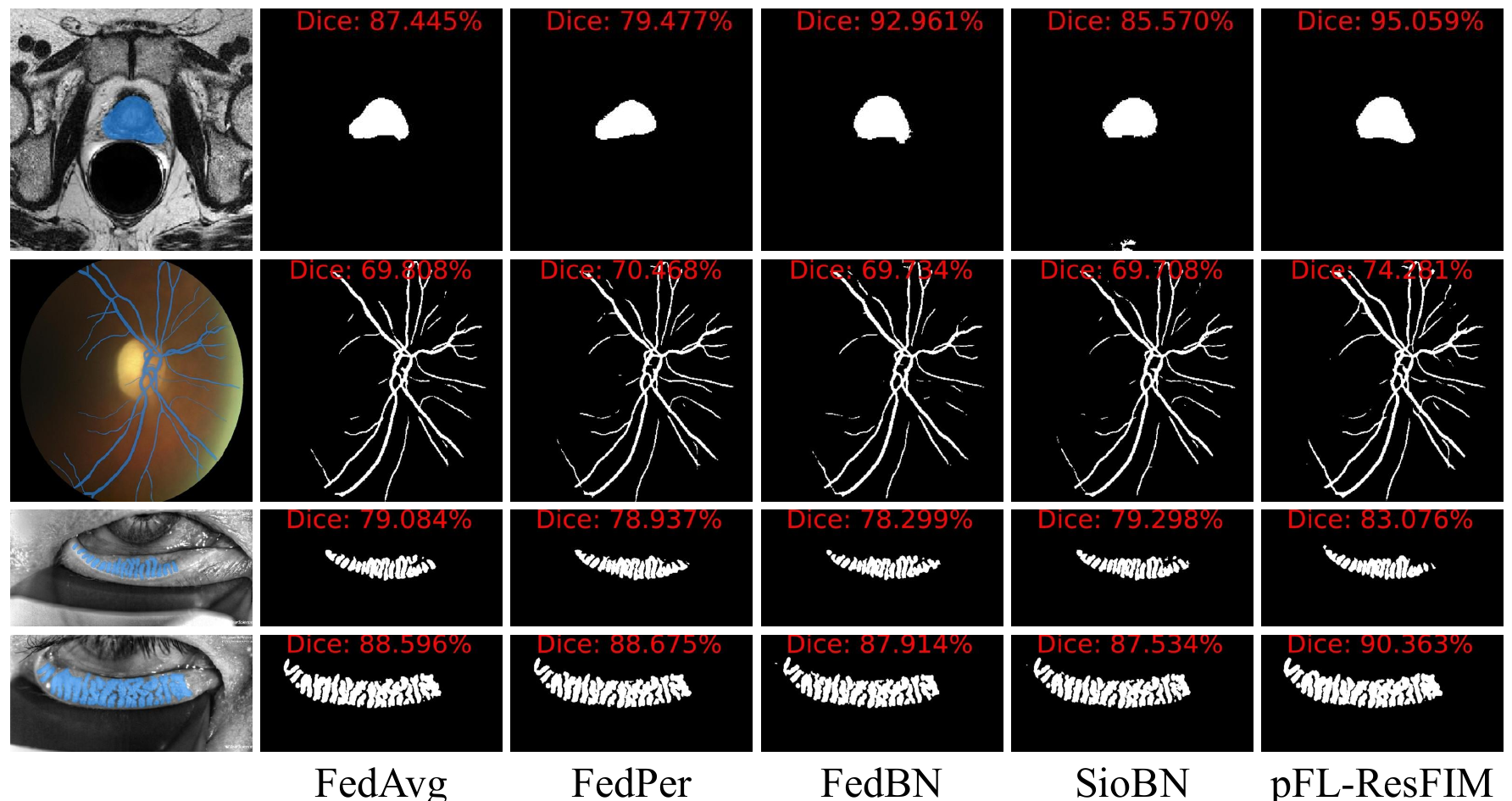}
    \vspace{-7mm}
    \caption{Visualization of segmentation results.}
    \label{fig:seg_results}
\end{figure}

\subsection{Comparison with State-of-the-art Methods}
We compare pFL-ResFIM with existing methods on three datastes, including FedAvg~\cite{fedavg}, FedBN~\cite{li2021fedbn}, SioBN~\cite{andreux2020siloed}, and FedPer~\cite{arivazhagan2019federated}. As shown in Tables~\ref{tab1}--\ref{tab3}, baseline methods~\cite{li2021fedbn, andreux2020siloed,arivazhagan2019federated} that designate certain layers as domain-sensitive often exhibit inconsistent performance across clients. For example, in Table~\ref{tab1}, FedBN achieves the highest Dice scores of 88.45\% and 87.77\% on the BMC and I2CVB sites, respectively, but underperforms on RUNMC and UCL. Similarly, in Table~\ref{tab2}, SioBN yields top performance on ORVS yet yields the lowest result on ChaseDB. These observations indicate that personalizing same layers fails to generalize across all clients. In contrast, pFL-ResFIM consistently achieves the best performance on nearly every client across all three datasets, demonstrating the effectiveness of its client-adaptive personalization strategy. We further provide qualitative comparisons by visualizing segmentation results of pFL-ResFIM and state-of-the-art methods in Fig.~\ref{fig:seg_results}. We can observe that pFL-ResFIM more accurately segment regular (prostate) or irregular (vessel and gland) objects than existing approaches, offering additional evidence of its practical advantages.

\begin{figure}
    \centering
    \includegraphics[width=\linewidth]{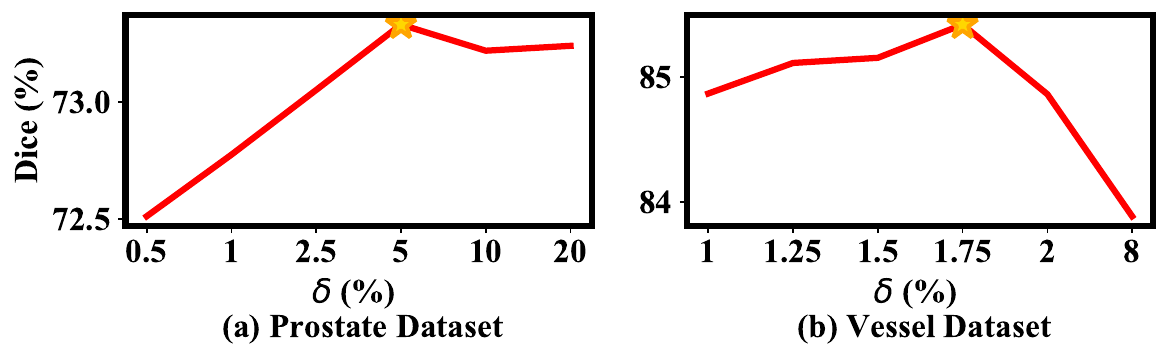}
    \vspace{-9mm}
    \caption{Performance of pFL-ResFIM with various $\delta\%$.}
    \label{fig:delta}
\end{figure}

\begin{figure}
    \centering
    \includegraphics[width=\linewidth]{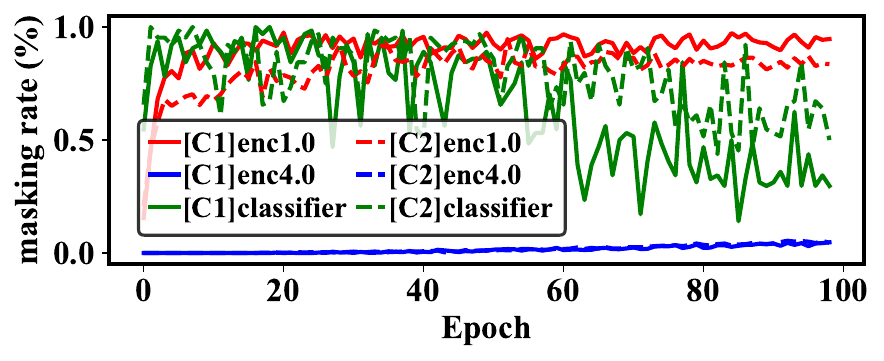}
    \vspace{-9mm}
    \caption{Masking rates of different layers of two clients.}
    \label{fig:maskingrate}
\end{figure}

\subsection{Ablation Study}
\textbf{Impact of Parameter $\delta\%$}. $\delta\%$ is a vital hyperparameter to determine how many parameters are personalized. Fig.~\ref{fig:delta} presents how the performance of pFL-ResFIM varies with different $\delta\%$. We observe a non-monotonic relationship between $\delta\%$ and model performance. The model's performance initially increases with $\delta\%$, peaks at an optimal value and then decreases on two datasets. The pattern suggests that maintaining an appropriate balance between personalized and shared parameters is essential for achieving optimal performance.

\noindent \textbf{Evaluation of ResFIM}. To evaluate the effectiveness of ResFIM on client-adaptive personalization, Fig.~\ref{fig:maskingrate} plots the masking rates of different layers of two client models during training on the vessel dataset. It can be observed that the input and classifier layers of two clients exhibit significantly different masking rates at the same epoch, whereas their intermediate layers (enc4.0) maintain relatively lower masking rates. This observation aligns with the current empirical consensus that both shallow and prediction layers are highly sensitive to domain shift. These results confirm the necessity of client-adaptive personalization and the effectiveness of ResFIM.

\section{Conclusion}
In this work, we introduce a new pFL framework called pFL-ResFIM that uses a Residual Fisher Information Matrix (ResFIM) based metric to measure the sensitivity of model parameters to domain shift. Based on the estimated ResFIM, we separate the parameters of a client model into domain-sensitive and domain-invariant parts and only aggregating the latter at the server for model personalization. 
Extensive experiments on public datasets demonstrate that pFL-ResFIM consistently outperforms state-of-the-art methods, validating its effectiveness for client-adaptive personalization.

\section{Compliance with Ethical Standards}
The use of open-access datasets complied with their licensing terms and did not require separate ethical approval. As for the in-house data, its use was approved by the Ethics Committee of the data provider's institution.

\bibliographystyle{IEEEbib}
\bibliography{refs}

\end{document}